\documentclass[11pt]{article}
\usepackage{histocrypt}
\usepackage{mathptmx}
\usepackage{url}
\usepackage{latexsym}
\usepackage{blindtext}
\usepackage{graphicx}
\usepackage{tabularx,booktabs}
\usepackage{multirow}

\usepackage{amsmath,amssymb}
\title{Learning to Decipher from Pixels — A Case Study of Copiale}

\author{ 
 Lei Kang \\
\textbf{Giuseppe De Gregorio} \\
 Computer Vision Center \\
 Universitat Autònoma de Barcelona \\
{\tt \{lkang, gdegregorio\}@cvc.uab.es} \\
 \And
Raphaela Heil \\
  Department of Linguistics\\
  Stockholm University, Sweden \\
  {\tt raphaela.heil@ling.su.se} \\
  \AND
\textbf{Alicia Fornés} \\
 Computer Vision Center \\
 Universitat Autònoma de Barcelona \\
 {\tt afornes@cvc.uab.es} \\
 \And
 Beáta Megyesi \\
  Department of Linguistics\\
  Stockholm University, Sweden \\
 {\tt beata.megyesi@ling.su.se}\\
}

\date{}

\begin{document}
\maketitle

\begin{abstract}
Historical encrypted manuscripts require both paleographic interpretation of cipher symbols and cryptanalytic recovery of plaintext. Most existing computational workflows rely on a transcription-first paradigm, in which handwritten symbols are transcribed prior to decipherment. This intermediate step is labor-intensive, error-prone, and not always aligned with the goal of direct plaintext recovery. We propose an end-to-end, transcription-free approach that directly maps handwritten cipher images to plaintext. Using the Copiale cipher as a case study, we introduce the first text-line-level dataset pairing cipher images with German plaintext. We show that pretraining on generic handwriting data followed by cipher-specific fine-tuning substantially improves decipherment accuracy. Our results demonstrate that transcription-free image-to-plaintext decipherment is both feasible and effective for historical substitution ciphers, offering a simplified and scalable alternative to traditional pipelines. \url{https://github.com/leitro/Decipher-from-Pixels-Copiale}.
\end{abstract}

\section{Introduction}

Historical encrypted manuscripts pose a dual challenge at the intersection of document analysis and cryptology~\cite{yin2019decipherment,megyesi2020decryption}. As handwritten artifacts with encoded content, they require both visual interpretation and cryptanalytic decoding. Consequently, most computational approaches adopt a transcription-first workflow in which symbols are transcribed into machine-readable ciphertext and then analyzed to recover plaintext~\cite{megyesi2020transcription,mendez2024structured}. This paradigm has shaped research practices in historical cryptology and digital humanities.

Despite its success, transcription-first processing presents substantial limitations. Historical ciphers often exhibit large and irregular symbol inventories, inconsistent spacing, and idiosyncratic writing conventions, making segmentation and transcription labor-intensive and error-prone. Errors propagate into downstream decipherment and require frequent reference to manuscript images. Moreover, ciphertext is rarely the final scholarly objective; researchers primarily seek readable plaintext, raising the question of whether explicit transcription is always necessary.

Early computational work assumed reliable ciphertext and focused on cryptanalytic modeling. Aldarrab introduced a probabilistic noisy-channel framework \cite{Aldarrab:2017} and explored early image-based decipherment using segmentation and clustering, identifying character segmentation as a major bottleneck. Yin et al. later formulated decipherment from manuscript images as an integrated task combining visual processing and statistical cryptanalysis ~\cite{yin2019decipherment}, demonstrating the feasibility of image-based approaches while retaining explicit symbol transcription. Together, these studies reflect a shift toward image-aware pipelines that nonetheless preserve staged processing.

\begin{figure*}[t]
  \centering
  \includegraphics[width=0.9999\linewidth]{imgs/arch_new.png}
  \caption{Overview of the training pipeline for our proposed Transcription-Free Decipherment paradigm. Stage I involves pretraining on a unified corpus of publicly available handwritten text-line datasets, followed by Stage II, where the model is fine-tuned on our curated Copiale image-to-plaintext text-line dataset.}
  \label{fig:arch}
\end{figure*}

Meanwhile, handwritten text recognition has advanced substantially through LSTM-based methods~\cite{bluche2017scan}, neural encoder--decoder models~\cite{kang2018convolve,kang2021candidate}, transformer-based architectures~\cite{kang2022pay}, and large-scale systems such as TrOCR~\cite{li2023trocr}. These developments enable direct image-to-text modeling and support end-to-end document understanding without explicit symbolic intermediates.

Building on these advances, we propose an end-to-end, transcription-free decipherment paradigm that directly maps handwritten cipher images to plaintext. Rather than supervising models with cipher symbols, we train them to generate decrypted natural-language text from pixel-level input. We evaluate this approach on the Copiale cipher~\cite{knight2011copiale,knight2012secrets}, a well-studied eighteenth-century German manuscript.

Our contributions are threefold. First, we release the first publicly available text-line-level dataset pairing Copiale cipher images with aligned German plaintext. Second, we demonstrate the feasibility of end-to-end image-to-plaintext decipherment for historical manuscripts. Third, we show that pretraining on generic handwriting data followed by cipher-specific fine-tuning substantially improves performance. These results indicate that transcription-free pipelines provide a scalable alternative to traditional multi-stage workflows in computational historical cryptology.

\section{Data}
\label{data}

\subsection{Handwriting Pretraining Data}
\label{sec:predata}

Pretraining is conducted on a merged corpus comprising 66,492 handwritten text lines drawn from four widely used and complementary benchmarks: IAM~\cite{marti2002iam}, CVL~\cite{kleber2013cvl}, RIMES~\cite{grosicki2011icdar}, and EU27~\cite{kohut2025practical}. This combination is designed to maximize diversity in writing styles, languages, scripts, and acquisition conditions, thereby improving the robustness and generalization of the learned representations.

The IAM dataset provides a large collection of English handwritten text lines produced by multiple writers under controlled scanning conditions, and is commonly used as a standard benchmark for offline handwriting recognition. CVL complements IAM by offering high-resolution handwritten documents from a distinct writer population, with variations in writing instruments and stroke dynamics that enrich intra-writer and inter-writer variability. RIMES contributes French handwritten text collected in a realistic mail-processing scenario, introducing linguistic diversity as well as challenges such as cursive writing, ligatures, and noise typical of real-world document workflows. Finally, EU27 extends coverage to a multilingual European setting, incorporating handwriting samples across multiple scripts and orthographic conventions, which is particularly valuable for learning script-agnostic and language-robust features.

The combined corpus is randomly split into training, validation, and test sets using an 80/10/10 ratio, while preserving the overall distribution of datasets and handwriting styles. Detailed statistics on text-line contributions from each dataset are reported in Table~\ref{tab:predata}.

\begin{table}[h!]
  \caption{Pretraining data for Stage I.}
  \label{tab:predata}
  \centering
  \resizebox{0.58\linewidth}{!}{
  \begin{tabularx}{0.6\linewidth}{lr}
    \toprule
    \textbf{Dataset} & \textbf{Textline Counts}\\
    \midrule
    CVL & 13,440\\
    IAM & 8,873\\
    RIMES & 12,104\\
    EU27 & 32,075\\
    \midrule
    Total & 66,492\\
    \bottomrule
  \end{tabularx}
  }
\end{table}

\subsection{Copiale Image-to-Plaintext Dataset}
\label{sec:copilaledata}

For cipher fine-tuning, we align handwritten Copiale text-line images with their corresponding German plaintext lines, producing the first dataset that supports direct image-to-plaintext decipherment of the manuscript. The number of lines, minimal and maximal number of characters and words per line for the training, validation and tests sets are described in Table \ref{tab:copiale}. The entire dataset is released publicly to support future research.


\begin{table}[h!]
  \caption{Fine-tuning data for Stage II: total number of lines, as well as the minimum and maximum number of characters and words per line, for the training, validation, and test sets of the Copiale.}
  \label{tab:copiale}
  \centering
  \resizebox{\linewidth}{!}{%
    \begin{tabular}{lrrrrr}
      \toprule
      \multirow{2}{*}{\textbf{Set}} 
        & \multirow{2}{*}{\textbf{Count}}
        & \multicolumn{2}{c}{\textbf{Char Length}} 
        & \multicolumn{2}{c}{\textbf{Word Length}} \\
      \cmidrule(lr){3-4} \cmidrule(lr){5-6}
        & & \textbf{Min} & \textbf{Max}
        & \textbf{Min} & \textbf{Max} \\
      \midrule
      Train & 1,269 & 1 & 67 & 1 & 14 \\
      Validation & 175 & 1 & 65 & 1 & 14 \\
      Test & 370 & 1 & 64 & 1 & 13 \\
      \bottomrule
    \end{tabular}
  }
\end{table}

\section{Method}

\subsection{Transcription-Free Decipherment Formulation}

We formulate decipherment as a sequence-to-sequence learning problem. In this paradigm, a model learns to transform one ordered sequence into another, without requiring explicit intermediate representations.
Given an input image $I$ representing a handwritten cipher text line, the model predicts a plaintext string $P$ in natural language. No explicit ciphertext representation is produced or supervised during training.

\subsection{Model Architecture}

We adopt TrOCR~\cite{li2023trocr}, a transformer-based encoder--decoder model for handwritten text recognition, and repurpose it for image-to-plaintext decipherment. A vision transformer encoder maps input images to latent visual embeddings, which are autoregressively decoded into plaintext tokens by a transformer decoder. Apart from adapting the output vocabulary, no architectural modifications are required. The overall training pipeline is shown in Figure~\ref{fig:arch}.

\subsection{Two-Stage Training Pipeline}

We adopt a two-stage training strategy consisting of handwriting pretraining followed by cipher-specific fine-tuning. In Stage~I, the model is pretrained on a unified corpus of publicly available handwritten text-line datasets (see Section~\ref{sec:predata}), to learn robust and largely style-invariant handwriting representations. In Stage~II, the pretrained model is fine-tuned on the Copiale image-to-plaintext dataset using German plaintext supervision (see Section~\ref{sec:copilaledata}). This strategy separates handwriting acquisition from task-specific learning: pretraining yields general visual representations independent of cipher symbols, while fine-tuning adapts these features for end-to-end decipherment.

\begin{table*}[t]
  \caption{End-to-end Copiale cipher decoding performance evaluated with CER (\%) and WER (\%), where lower is better.}
  \label{tab:sota}
  \centering
  \resizebox{0.8\linewidth}{!}{
  \begin{tabularx}{0.86\textwidth}{lcccccccc}
    \toprule
    \multirow{2}{*}{\textbf{Method}} & \multirow{2}{*}{\textbf{Stage I}} & \multirow{2}{*}{\textbf{Stage II}}
      & \multicolumn{2}{c}{\textbf{Train Set}} 
      & \multicolumn{2}{c}{\textbf{Validation Set}}
      & \multicolumn{2}{c}{\textbf{Test Set}} \\
    \cmidrule(lr){4-5} \cmidrule(lr){6-7} \cmidrule(lr){8-9}
      & & & \textbf{CER↓} & \textbf{WER↓}
      & \textbf{CER↓} & \textbf{WER↓}
      & \textbf{CER↓} & \textbf{WER↓}  \\
    \midrule
    Baseline & $-$ & \checkmark & 42.58 & 93.06 & 45.85 & 101.81 & 46.10 & 98.48\\
    \midrule
    \textbf{Ours} & \checkmark & \checkmark & \textbf{0.19} & \textbf{0.28} & \textbf{11.02} & \textbf{34.14} & \textbf{11.03} & \textbf{33.03}\\
    \bottomrule
  \end{tabularx}
  }
\end{table*}

\begin{figure*}[t]
  \centering
  \includegraphics[width=0.9\linewidth]{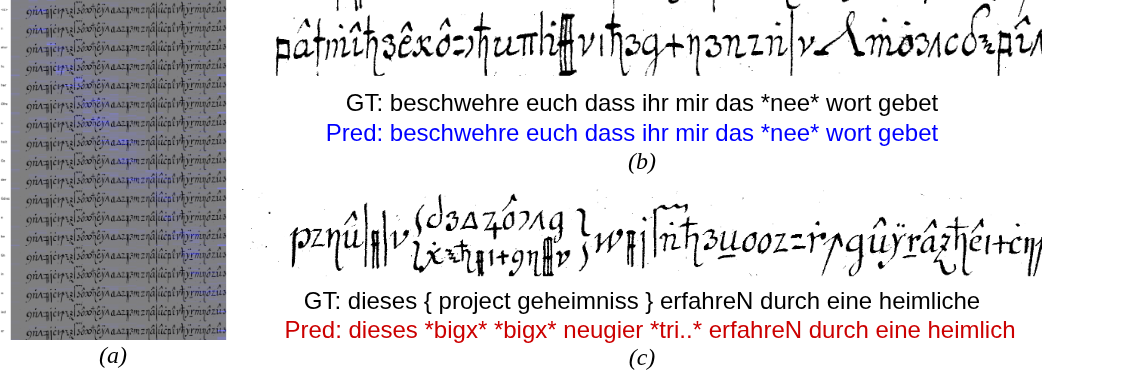}
\caption{(a) Attention visualization illustrating alignment between handwritten cipher regions and decoded plaintext tokens. (b) Successful prediction example, where the model output (blue) matches the ground truth (black). (c) Failure case, where predictions (red) deviate from the ground truth (black), illustrating typical error patterns.}
  \label{fig:res_pos_neg}
\end{figure*}

\section{Experiments}

\subsection{Implementation Details}
In Stage~I pretraining, the model is trained for 5 epochs with a learning rate of $5\times10^{-5}$ and a batch size of 64. For Stage~II, we adopt a learning rate of $2\times10^{-5}$ and a batch size of 64, and apply early stopping with a patience of 20 epochs based on validation performance, resulting in a total of 21 training epochs. 
AdamW is used as the optimizer, and the backbone model is \texttt{microsoft/trocr-base-handwritten}. All experiments are conducted on a single NVIDIA 4090 GPU.

\subsection{Evaluation Metrics}

We report Character Error Rate (CER) and Word Error Rate (WER), which are standard metrics for handwriting recognition and sequence prediction. CER is defined as the normalized Levenshtein distance at the character level, $\mathrm{CER} = \frac{S_c + D_c + I_c}{N_c}$, where $S_c$, $D_c$, and $I_c$ denote the numbers of character substitutions, deletions, and insertions, respectively, and $N_c$ is the total number of characters in the reference text. Similarly, WER measures error at the word level and is defined as $\mathrm{WER} = \frac{S_w + D_w + I_w}{N_w}$, where $S_w$, $D_w$, and $I_w$ denote the numbers of word substitutions, deletions, and insertions, and $N_w$ is the total number of words in the reference text.

\subsection{Results}
The results are shown in Table \ref{tab:sota}. Handwriting pretraining yields a CER of 5.93\% and WER of 17.99\% on held-out handwriting data, indicating strong general visual representations. Table~\ref{tab:sota} compares direct fine-tuning on Copiale against our two-stage approach. The pretrained model dramatically outperforms the baseline, reducing test-set CER from 46.10\% to 11.03\% and WER from 98.48\% to 33.03\%. These results show that non-cipher handwriting data substantially improves decipherment accuracy, even though it contains no cryptographic structure.

\section{Qualitative Analysis}

Qualitative evaluation indicates that the model learns semantically meaningful alignments between handwritten cipher symbols and corresponding plaintext segments, even without explicit ciphertext-level supervision. As shown in Figure~\ref{fig:res_pos_neg}(a), the attention maps reveal coherent correspondences between spatial regions of the cipher input and decoded plaintext tokens, suggesting that the model captures the underlying structure of the cipher.

Figures~\ref{fig:res_pos_neg}(b) and~(c) present representative successful and failure cases, respectively. In Figure~\ref{fig:res_pos_neg}(b), the predicted plaintext closely matches the ground truth, demonstrating reliable decoding performance across diverse inputs. In contrast, Figure~\ref{fig:res_pos_neg}(c) illustrates typical failure patterns, where predictions deviate from the ground truth, highlighting the limitations of the current model.

\section{Conclusion}

We proposed Transcription-Free Decipherment, an end-to-end approach to historical cipher decipherment that directly maps handwritten cipher images to plaintext. Using the Copiale cipher, we introduced a new publicly available image-to-plaintext dataset and demonstrated that handwriting pretraining on non-cipher data dramatically improves decipherment performance. By collapsing transcription and decipherment into a single learnable mapping, our approach reduces annotation effort, simplifies processing pipelines, and opens new directions for scalable analysis of historical encrypted manuscripts.


\section*{Acknowledgments}
This work has been supported by Riksbankens Jubileumsfond, grant M24-0028: Echoes of History: Analysis and Decipherment of Historical Writings (DESCRYPT); the Beatriu de Pinós del Departament de Recerca i Universitats de la Generalitat de Catalunya (2022 BP 00256); European Lighthouse on Safe and Secure AI (ELSA) from the European Union’s Horizon Europe programme under grant agreement No 101070617; the Spanish projects CNS2022-135947 (DOLORES), PID2021-126808OB-I00 (GRAIL) and PID2024-157778OB-I00 (SUKIDI), the Consolidated Research Group 2021 SGR 01559 from the Research and University Department of the Catalan Government, and PID2023-146426NB-100 funded by MCIU/AEI/10.13039/501100011033 and FSE+. Alicia Fornés acknowledges financial support for her general research activities from ICREA under the ICREA Academia (Departament de Recerca i Universitats de la Generalitat de Catalunya).


\bibliographystyle{histocrypt}
\bibliography{histocrypt}

\end{document}